\documentclass[conference]{IEEEtran}
\IEEEoverridecommandlockouts
\usepackage{cite}
\usepackage{amsmath,amssymb,amsfonts}
\usepackage{graphicx}
\usepackage{textcomp}
\usepackage[table]{xcolor}
\usepackage{booktabs}
\usepackage{subcaption}
\usepackage{caption}
\usepackage{float}
\usepackage{lipsum} 
\usepackage{hyperref} 
\usepackage{tikz}
\usepackage{listings}
\usepackage[ruled,vlined]{algorithm2e}
\usepackage{multicol}
\usepackage{multirow}

\newcommand\copyrighttext{%
  \footnotesize \textcopyright \the\year{} IEEE. Personal use of this material is permitted. Permission from IEEE must be obtained for all other uses, including reprinting/republishing this material for advertising or promotional purposes, collecting new collected works for resale or redistribution to servers or lists, or reuse of any copyrighted component of this work in other works.}
  
\newcommand\copyrightnotice{%
\begin{tikzpicture}[remember picture,overlay]
\node[anchor=south,yshift=10pt] at (current page.south) {\fbox{\parbox{\dimexpr0.75\textwidth-\fboxsep-\fboxrule\relax}{\copyrighttext}}};
\end{tikzpicture}%
}

\graphicspath{ {./images/} }
\def\BibTeX{{\rm B\kern-.05em{\sc i\kern-.025em b}\kern-.08em
    T\kern-.1667em\lower.7ex\hbox{E}\kern-.125emX}}
\begin{document}

\title{Diverse Prompts: Illuminating the Prompt Space of Large Language Models with MAP-Elites
}

\author{\IEEEauthorblockN{ Gabriel Machado Santos}
\IEEEauthorblockA{\textit{Federal University of Uberlândia} \\
\textit{Computer Science Faculty}\\
São Paulo, Brazil \\
gabrielmsantos@gmail.com}
\and
\IEEEauthorblockN{Rita Maria da Silva Julia}
\IEEEauthorblockA{\textit{Federal University of Uberlândia} \\
\textit{Computer Science Faculty}\\
Uberlândia, Brazil \\
rita@ufu.br}
\and
\IEEEauthorblockN{Marcelo Zanchetta do Nascimento}
\IEEEauthorblockA{\textit{Federal University of Uberlândia} \\
\textit{Computer Science Faculty}\\
Uberlândia, Brazil \\
marcelo.nascimento@ufu.br}
}

\bibliographystyle{./IEEEtran}
\maketitle

\copyrightnotice

\begin{abstract}
Prompt engineering is essential for optimizing large language models (LLMs), yet the link between prompt structures and task performance remains underexplored. This work introduces an evolutionary approach that combines context-free grammar (CFG) with the MAP-Elites algorithm to systematically explore the prompt space. Our method prioritizes quality and diversity, generating high-performing and structurally varied prompts while analyzing their alignment with diverse tasks by varying traits such as the number of examples (shots) and reasoning depth. By systematically mapping the phenotypic space, we reveal how structural variations influence LLM performance, offering actionable insights for task-specific and adaptable prompt design. Evaluated on seven BigBench Lite tasks across multiple LLMs, our results underscore the critical interplay of quality and diversity, advancing the effectiveness and versatility of LLMs.
\end{abstract}


\section{Introduction}

%The rapid advancement of large language models (LLMs), such as ChatGPT based on the GPT architecture, has revolutionized natural language processing (NLP) \cite{raiaan2024review}. These models exhibit exceptional abilities across various domains, highlighting the growing importance of prompt engineering. By optimizing model performance through effective prompts, prompt engineering bridges the gap between user intent and model output, making it an essential tool for diverse applications, from code intelligence to customer support \cite{schulhoff2024prompt, wang2023prompt}.

%Despite its significance, understanding how structural factors like examples, reasoning depth, and context impact the performance of LLMs still remains underexplored. These \emph{phenotypic characteristics} of prompts are crucial for uncovering diverse paths to high-quality prompts, particularly in zero-shot and low-resource settings, where traditional data-intensive approaches are impractical \cite{wen2024hard}. Effective prompts help LLMs understand nuanced contexts, enabling them to adapt to a wide range of tasks \cite{wang2023prompt, ekin2023prompt}

The rapid advancement of Generative Pre-Trained Transformer (GPT)-based Large Language Models (LLMs), such as ChatGPT, has revolutionized the field of Natural Language Processing (NLP) \cite{raiaan2024review}. These models excel across domains, emphasizing the importance of prompt engineering to optimize performance and bridge user intent with model output \cite{schulhoff2024prompt, wang2023prompt}.

However, the impact of structural factors like examples, reasoning depth, and context on LLM performance remains underexplored. Understanding these \emph{phenotypic characteristics} is vital for identifying diverse, high-quality prompts, particularly in zero-shot and low-resource settings where traditional approaches are less effective \cite{wen2024hard, ekin2023prompt}.

%In this work, we propose an evolutionary approach to exploring the prompt space by using a context-free grammar (CFG) to systematically generate prompt structures, as well as the MAP-Elites \cite{mouret2015illuminating} algorithm to ensure diversity. By leveraging Quality Diversity (QD) algorithms, we aim to discover diverse pathways to high-quality outputs and examine how these structural variations influence LLM performance across various tasks and phenotypic dimensions.

In this work, we propose a novel evolutionary approach to prompt engineering, combining context-free grammar (CFG) with the MAP-Elites algorithm \cite{mouret2015illuminating}  to explore the prompt space systematically. Our methodology emphasizes both quality and diversity, uncovering high-performing and structurally varied prompts. By varying phenotypic traits, such as the number of examples (shots) and reasoning depth, we analyze how prompt structures align with task-specific requirements.

Unlike previous works that focus primarily on improving prompt design for specific tasks \cite{saletta2024exploring}, our approach extends beyond optimization to also examine how different prompts align with a variety of tasks. By employing MAP-Elites, we systematically explore the phenotypic space, identifying broad design principles that enhance both performance and adaptability. This dual focus on optimization and alignment positions our work as a significant contribution to prompt engineering, offering a comprehensive framework for creating and analyzing effective prompts across diverse scenarios.

This work aims to characterize the relationship between structural prompt variations and LLM performance within BigBench tasks \cite{bigbenchlite}. Specifically, we focus on: (1) expanding the diversity of prompt designs using the MAP-Elites algorithm to systematically explore viable structures beyond random or naive methods, (2) analyzing phenotypic characteristics like examples, reasoning depth, word count, and context inclusion to evaluate their impact on task performance, and (3) examining whether prompt structures generalize across tasks or require task-specific designs.

This way, the main contributions of the present work are the following: (1) a novel combination of grammar-based prompt generation and MAP-Elites for quality-diversity exploration, enabling precise control over key phenotypic traits while identifying diverse, high-performing prompts; (2) a comprehensive evaluation across BigBench tasks, elucidating how prompt structures influence task-specific outcomes; and (3) insights into prompt generalization and specialization, providing practical guidance for designing prompts adaptable to a range of real-world applications. This work advances prompt engineering by systematically exploring diverse designs and offering actionable insights for NLP practitioners.

%% Outline
%The remainder of this paper is organized as follows: Section~\ref{sec:related_work} reviews related work on LLMs and prompt engineering, quality-diversity methods, and automatic prompt generation. Section~\ref{sec:methodology} details the proposed framework, including our context-free grammar and MAP-Elites setup. Section~\ref{sec:experiments} outlines the experimental design, datasets, and metrics. Section~\ref{sec:results-discussion} presents results, analyses, and discussions on findings. Finally, Section~\ref{sec:conclusion} summarizes contributions, limitations, and future directions, highlighting the broader impact on prompt engineering and LLM research.
The remainder of the paper is organized as follows: Section~\ref{sec:related_work} reviews related work; Section~\ref{sec:methodology} describes our framework and MAP-Elites setup; Section~\ref{sec:experiments} covers the experimental design; Section~\ref{sec:results-discussion} discusses results and findings; and Section~\ref{sec:conclusion} concludes with contributions, limitations, and future directions.

\section{Background and Related Work}
\label{sec:related_work}

\subsection{Large Language Models and Prompt Engineering}

Over the past decade, LLMs have transformed NLP and machine learning (ML), progressing from static embeddings like Word2Vec \cite{mikolov2013distributed} and GloVe \cite{pennington2014glove} to contextual models such as BERT \cite{devlin2018bert}, GPT \cite{radford2018improving}, and Transformer architectures \cite{vaswani2017attention}. Modern LLMs, including GPT-3 \cite{brown2020language} and LLaMA \cite{touvron2023llama}, handle complex tasks using massive parameters and hardware innovations \cite{shoeybi2019megatron}, but remain sensitive to task formats, underscoring the importance of prompt engineering.

Prompt engineering involves crafting effective textual inputs to optimize LLM outputs \cite{liu2023pre}. Techniques like zero-shot, few-shot \cite{min2022rethinking}, chain-of-thought prompts \cite{wei2022chain}, and role context, for guiding style and content \cite{mazare2018training},  enhance performance while reducing the need for fine-tuning \cite{schick2020exploiting}. This approach is especially valuable for low-resource scenarios, making LLMs more adaptable and accessible \cite{wang2023prompt, chang2024survey}.

\subsection{Quality Diversity Algorithms}

%Quality-Diversity algorithms (QD) are a class of evolutionary techniques designed to discover diverse, high-performing solutions \cite{pugh2016quality, cully2015robots, mouret2015illuminating}. Unlike traditional optimization, which targets a single global optimum, QD methods aim to explore the entire solution space, mapping high-quality and diverse candidates.

%Algorithms like MAP-Elites \cite{mouret2015illuminating} optimize performance while exploring variations in \emph{behavioral descriptors} or \emph{phenotypic characteristics}, yielding solutions that are both effective and structurally diverse. This is particularly useful in domains where multiple solutions hold value or feature interactions are complex.

%QD algorithms have proven versatile across robotics \cite{cully2015robots}, reinforcement learning \cite{stanton2018deep}, and creative computing \cite{nguyen2016understanding}. Their ability to expose diverse solution spaces makes them ideal for exploring prompt structures, uncovering both high-performing and distinct designs.

Quality-Diversity (QD) algorithms are evolutionary techniques that discover diverse, high-performing solutions by exploring the entire solution space, rather than targeting a single global optimum \cite{pugh2016quality, mouret2015illuminating}. Algorithms like MAP-Elites \cite{mouret2015illuminating} optimize performance while varying \emph{behavioral descriptors} or \emph{phenotypic characteristics}, yielding effective and structurally diverse solutions. Widely applied in robotics \cite{cully2015robots}, reinforcement learning \cite{stanton2018deep}, and creative computing \cite{nguyen2016understanding}, QD algorithms are particularly suited for exploring diverse prompt structures to uncover both high-performing and distinct designs.

\subsection{Automatic Prompt Generation in Literature}

The rise of LLMs has highlighted the need for automated prompt generation to replace time-consuming, inconsistent manual engineering. Methods include gradient-based approaches like prefix-tuning \cite{li2021prefix}, search-based strategies such as AutoPrompt \cite{shin2020autoprompt}, and frameworks for modular prompt composition using reusable blocks or CFGs \cite{raffel2020exploring, wang2024grammar}.

While these methods focus on finding task-specific, high-performing prompts, our work uses MAP-Elites to explore the structural diversity of prompts, uncovering a broader range of effective designs that may generalize across tasks. Additionally, our approach establishes a framework for analyzing task sensitivity to specific prompt structures, providing deeper insights into how different designs influence performance across varied scenarios.

\section{Methodology}
\label{sec:methodology}

This section describes the proposed methodology to study the efficacy of prompt diversity applied to distinct tasks. 

%The first step in the method is to structure a generic, dynamic, and flexible way to generate prompts so that we can explore diversity in prompts without having to explicitly or manually set it. For this task we chose to represent the prompt structure with a CFG. 

\subsection{Context-Free Grammar for representing genotypes}

To systematically generate diverse prompt structures, we propose a Context-Free Grammar \cite{chomsky2014aspects}. A CFG is defined by a four-tuple 
\[
G = (V, \Sigma, R, \textbf{P}),
\]
%where \(V\)  is a finite set of nonterminal symbols, \(\Sigma\) is a finite set of terminal symbols, \(R\) is a set of production rules of the form $A \rightarrow \alpha$, where $A$ \in \(V\)  is a nonterminal and $\alpha$ is a sequence of terminals and/or nonterminals (possibly empty), and \textbf{P} \in \(V\) is the start symbol (or axiom).
where \(V\) is a finite set of nonterminal symbols, \(\Sigma\) is a finite set of terminal symbols, \(R\) is a set of production rules of the form \(A \rightarrow \alpha\), where \(A \in V\) is a nonterminal and \(\alpha\) is a sequence of terminals and/or nonterminals (possibly empty), and \(\mathbf{P} \in V\) is the start symbol (or axiom).

In our setup,  $V = \{\textbf{P}, S, E, I, C, T, R, X, N\}$  represents the set of variables, and $ \Sigma = \{``*"\} $ is a set of terminal elements, where  $``*"$ represents a string. Strings may include generic elements, denoted by double parentheses (()), or task-specific elements, denoted by double square brackets [[*]]. When instantiating a prompt, these placeholders are replaced with strings from either a generic table or task-specific context.

The rules $R$ are represented as follows:

%In our setup the nonterminals include symbols such as \(\langle \text{prompt}\rangle\), \(\langle \text{base}\rangle\), \(\langle \text{nshot}\rangle\), and so forth, whereas terminals consist of literal strings (e.g., “\texttt{Consider this example:}”) and placeholders for task-specific elements (e.g., \([\![\text{EXAMPLE}]\!]\)).

\begin{align*}
R:
\textbf{P} &\rightarrow STEI \ | \ SEI \ | \ CSTEI \ | \ CSEI \\
%B &\rightarrow SD \ | \ S \\
S &\rightarrow RX \ | \ R\\
X &\rightarrow \text{``Consider this example: [[example]]''} \\
X &\rightarrow \text{``Consider these examples: [[example]],''} N \\
N &\rightarrow \text{``[[example]]''} \ | \ \text{``[[example]]''} N \\
E &\rightarrow \text{``[[task entry]]''} \\
I &\rightarrow \text{``[[LLM instruction]]''} \\
R &\rightarrow \text{``[[task request]]''} \\
C &\rightarrow \text{``((c1))''} \ | \ \text{``((c2))''} \ | \ \dots \ | \ \text{``((c10))''} \\
T &\rightarrow \text{``((t1))''} \ | \ \text{``((t2))''} \ | \ \dots \ | \ \text{``((t10))''} \\
\end{align*}

%In our CFG, the symbol $\langle \text{prompt}\rangle$ is the entry point, allowing either a simple prompt ($\langle \text{base}\rangle\ \langle \text{input}\rangle\ \langle \text{instruction}\rangle$) or one that includes $\langle \text{context}\rangle$ for role-based guidance. The base of the prompt can either include $\langle \text{nshot}\rangle\ \langle \text{nsbs}\rangle$—representing few-shot examples and chain-of-thought depth—or just $\langle \text{nshot}\rangle$. The symbol $\langle \text{nshot}\rangle$ in turn can come with or without concrete examples, which are retrieved from the dataset ($\langle \text{examples}\rangle$). Each example may be expanded into a single instance or multiple instances via recursion in $\langle \text{example}\rangle$. The remaining elements ($\langle \text{input}\rangle, \langle \text{instruction}\rangle, \langle \text{request}\rangle, \langle \text{context}\rangle, \langle \text{nsbs}\rangle$, and $\langle \text{number}\rangle$) serve as terminals, filling in specific task-related text (e.g., user queries, instructions, or requests), context roles (e.g., “C1,” “C2,” … “C10”), or step indicators for chain-of-thought prompts (“COT1,” “COT2,” … “COT20”). This modularity ensures maximum flexibility in prompt design.

Next, a brief explanation about the non terminals that make up $V$. The nonterminal $S$ (shots) comprises the examples included in the prompt. The symbols $C$ (context-role) and $T$ (thoughts) represent generic elements for the context-role and reasoning depth, respectively. Each was empirically instantiated with 10 variations and used consistently throughout the study. $C$ provides role-based guidance, such as “You are a researcher presenting your findings at a scientific conference, answering questions from fellow scientists.”, and T specifies reasoning depth with requests like “Think step by step. The number of steps you must consider is $<$number$>$.”, where $<$number$>$ ranges from 1 to 10.

The nonterminal symbols $X$ (examples), $N$ (next example), $E$ (task entry), $I$ (LLM instruction), and $R$ (task request) correspond to task-specific strings. $E$ is a random entry from the task dataset, $R$ defines the task request (e.g., “Please answer the following questions about which words certain pronouns refer to.”) for the Winowhy task, whereas $I$ provides instructions on how the LLM should respond. Examples ($X$) are also task-specific, offering questions and answers to guide the LLM’s response.

For simplicity, we implement the CFG expansion via a depth-first traversal of the production rules. During the generation process, each nonterminal symbol is expanded into one of its possible productions, as dictated by a list of rule indices. For example, we encode a particular expansion path as

\[
\begin{aligned}
Gen: [\,&\textbf{P}2, C6, 
      S0, R0,
      X0, 
      T3, E0, 
      I0
\,],
\end{aligned}
\]
where each pair $\langle \text{nonterminal}, i\rangle\,$ denotes which production rule $i$ is chosen for that nonterminal. Here, $Gen$ operates as the genotype in our evolutionary framework, encoding the structure of the prompt. The grammar is processed in a tree-like way, where the nonterminals are expanded until only terminals remain. The generated genotype will then instantiate a prompt to generate an adequate input for the LLM. Such a fully specified prompt incorporates actual text and, depending on the type of the intended input, it also comprises task-specific tokens referring to context-roles and/or reasoning thoughts.

By selectively including or excluding the nonterminal symbols $C$, $X$ and $T$, our CFG supports multiple prompt configurations. These include zero-shot prompts, which contain no examples; few-shot prompts, which include one or two examples; and many-shot prompts, with more than two examples. Prompts can also be enhanced with context by incorporating a context-role element ($C$), or with chain-of-thought, specifying the depth of reasoning ($T$) to guide step-by-step responses. 

Building on this, we use MAP-Elites to optimize diversity and quality, as detailed next.

\subsection{MAP-Elites for leveraging prompt diversity}
\label{sec:MAP-Elites}

To further explore how prompt diversity influences the performance of LLMs, we employ an adapted version of the MAP-Elites algorithm. By varying the phenotypic characteristics of the prompts — such as length, reasoning depth, and number of examples — ,  a set of candidate solutions is created and enhanced through an evolutionary process whose purpose, at each iteration, is to improve the fitness os such solutions (which are stored into a structured archive), balancing exploration of the phenotypic space with high-performance objectives. This method provides a robust framework for analyzing the relationship between prompt design and LLM behavior. Below, we detail the steps of the algorithm and its integration with the CFG-generated prompts. 
{\small
\begin{algorithm}
\caption{MAP-Elites for Prompt Generation}
\SetKwInOut{Input}{Input}
\SetKwInOut{Output}{Output}
\Input{CFG (Context-Free Grammar), Task Dataset $T$, $population\_size$, $num\_iterations$, $Evaluator$ (LLM API), $bin\_sizes$, $mut\_rate$, $mut\_chance$, $num\_evaluations$}
\Output{Archive $A$ with diverse, high-performing prompts}

Initialize an empty archive $A$\;
Generate initial population $P$ of $population\_size$ individuals by processing the CFG\;
\For{$t = 1$ \KwTo $num\_iterations$}{
  \ForEach{$individual \in P$}{
    Compute fitness by running the genotype on $num\_evaluations$ task instances using the $Evaluator$\;
    Compute the phenotypes of the individual\;
    Assign the individual to a bin based on $bin\_sizes$\;
    Add or replace the individual in archive $A$ if it improves fitness\;
  }
  Generate offspring by mutating $mut\_rate$\% of $P$ with mutation probability $mut\_chance$\%\;
  Update $P$ with new individuals from $A$\;
}
\Return{$A$}
\end{algorithm}
}

The adapted MAP-Elites algorithm begins by initializing an empty archive  $A$, which serves as the repository for diverse and high-performing prompts. The initial population $P$, consisting of $population\_size$ individuals (genotypes), is created randomly based on the CFG rules.

The algorithm then iterates over $num\_iterations$, refining the population and archive with each iteration. For every individual in the population, its fitness is computed by running the corresponding genotype on a set of $num\_evaluations$ task instances from the given dataset $T$, using the provided $Evaluator$ (an interface for LLM APIs) for evaluation. The fitness $f$ is determined by the proportion of correctly solved tasks \ref{eq:fitness}, making it a measure of the effectiveness of the prompt.

% FORMULA HERE
\begin{equation}
\label{eq:fitness}
    f = \frac{{num\_correct}}{num\_evaluations}
\end{equation}
where $num\_correct$ is the number of task instances the prompt solved correctly, and $num\_evaluations$ is the total number of task instances evaluated.  

After calculating an individual’s fitness, its phenotypic features—such as prompt length, reasoning depth, number of examples (shots), and context-role—are determined. The individual is placed into a bin, a category within the archive A defined by the bin\_sizes parameter, which groups individuals with similar features. If the new individual outperforms the current occupant of its bin, it replaces it in such bin.

%In place of crossover operations, which are often used in evolutionary algorithms, this implementation focuses on mutation for simplicity. Here, mutations involve altering the genotype of $mutation_rate$ percentage of random selected individuals from the current population, with $mutation_chance$ percentage of chance of adjusting properties like the number of shots, depth of reasoning (e.g., adding or removing chain-of-thought steps), or contextual information. These mutations ensure that the algorithm explores a wide range of potential solutions while maintaining the diversity of the population.

For simplicity, this implementation replaces crossover with mutation. Mutations are applied to  $mut\_rate$ \% of randomly selected individuals, each with a  $mut\_chance$ \% probability of modifying properties such as the number of shots, reasoning depth (e.g., varying the placeholder $<$number$>$ in the symbol "$T$" in the CFG), or contextual information.

After mutation, the population is updated by incorporating individuals from the archive $A$, forming the basis for the next iteration. This iterative process continues until the specified number of iterations is reached. At the end of the algorithm, the archive $A$ contains a diverse set of high-performing prompts, optimized for both quality and phenotypic diversity. This adaptation of MAP-Elites emphasizes exploration over exploitation, making it a robust method for analyzing the relationship between prompt diversity and the performance of large language models.

\section{Experimental Setup}
\label{sec:experiments}

This section outlines the experimental setup to evaluate the proposed methodology for automatic prompt generation, as well as its impact on LLM performance. The experiments aim to answer three key questions:
	1.	Can the MAP-Elites algorithm enhance prompt diversity compared to random sampling?
	2.	How do prompt structures (e.g., examples, reasoning depth) correlate with LLM accuracy?
	%3.	Do certain prompt structures generalize across tasks, or are task-specific designs required?
        3. Are there prompt structures that can be adapted to general tasks, or is it better to create specific prompts for each type of task?

%Seven tasks from BigBench Lite (BBL) \cite{bigbenchlite} \cite{srivastava2022beyond} were evaluated across four LLMs. For the sake of transparency and reproducibility, detailed descriptions of tasks and models, as well as the code and results, are available on GitHub\footnote{\href{https://github.com/gabrielmsantos/diverse-prompts}{https://github.com/gabrielmsantos/diverse-prompts}}.

Seven tasks from BigBench Lite (BBL) \cite{bigbenchlite} \cite{srivastava2022beyond} were evaluated across four LLMs. For the sake of transparency and reproducibility, detailed descriptions of tasks and models, as well as the code and results, are publicly available\footnote{\href{https://doi.org/10.5281/zenodo.14630714}{https://doi.org/10.5281/zenodo.14630714}}.

\subsection{Task Datasets}

We used seven tasks from the BBL benchmark \cite{srivastava2022beyond, saletta2024exploring }, a curated subset of BigBench designed to balance computational feasibility with task complexity. These tasks evaluate diverse reasoning and comprehension abilities:
\begin{itemize}
    \item Formal Fallacies and Syllogisms Negation (FFSN) tests formal logic by identifying invalid syllogisms.
    \item Known Unknowns (KU) evaluates factual reasoning by distinguishing verifiable from unverifiable claims.
    \item Logical Deduction 3 (LD3) focuses on multi-step logical deductions.
    \item Play Dialog Same or Different (PDSD) assesses conversational comprehension via semantic equivalence in dialogs.
    \item Strange Stories Boolean (SSB) tests boolean logic in contextual narratives.
    \item StrategyQA (SQA) requires multi-step reasoning for open-ended questions.
    \item Winowhy evaluates ambiguity resolution in contextual scenarios.
    
\end{itemize}

Most tasks have binary answers, except for LD3, which features ternary ones. This diverse set ensures comprehensive evaluation across linguistic and cognitive challenges.

\subsection{Language Models}

%The proposed methodology was evaluated using four state-of-the-art LLMs, selected for their diverse architectures and design objectives. Models smaller than 10 billion parameters were chosen due to cost constraints and the lack of infrastructure required to efficiently run larger models locally. Each model was accessed through Hugging Face Inference Endpoints \cite{huggingfaceendpoints} using REST API calls. Below is a brief description of each model:

The proposed methodology was evaluated on four state-of-the-art LLMs, chosen for their diverse architectures and parameter sizes (under 10B) due to cost and infrastructure constraints. Models were accessed via Hugging Face Inference Endpoints \cite{huggingfaceendpoints} through REST API calls. Descriptions of each model are provided below:

\begin{itemize}
    \item Starling-LM-7B-Alpha-BII \cite{starling7b}: A 7-billion parameter model designed for general-purpose instruction following.
    \item LLaMA-3-1-8B-Instruct-LVT \cite{llama31}: An 8-billion parameter model fine-tuned for structured instructions and reasoning. 
    \item Phi-3-5-Mini-Instruct-IVR \cite{phi35}: A compact 3.5-billion parameter model optimized for computational efficiency while maintaining strong reasoning capabilities.
    \item Qwen2-5-7B-Instruct-MLN \cite{qwen25}: A 7-billion parameter model tailored for multi-lingual instruction tasks.
\end{itemize}

To enhance efficiency and minimize latency during evaluation, the output length of each model was restricted to a maximum of three tokens. This constraint ensured that only the essential answer was returned, excluding additional explanations or verbose outputs. 

\subsection{Baseline and Parameters}

To evaluate the MAP-Elites algorithm, we used random sampling as a baseline for diversity. The algorithm ran with a population of 50 individuals per generation for 10 iterations. Both the mutation rate (mut\_rate) and mutation probability (mut\_chance) were set to 40\%. The fitness of each individual was evaluated on 50 task instances, with bin sizes of (2, 25, 2), discretizing the number of examples, prompt length (in words), and reasoning depth. This setup balanced phenotypic exploration with computational efficiency.

Tests were carried out for each LLM using both MAP-Elites and Random Search algorithms, with the LLM temperature set to 0. This resulted in a total of 56 runs (one per algorithm for each of the four LLMs across seven datasets), where the duration of each run was,on average, 7 hours.

All experiments were conducted on a MacBook Pro with an Apple M3 Max chip and 48 GB of memory. This hardware provided the necessary computational power to handle prompt generation, LLM API calls, and result processing with efficiency and reliability.

%\section*{Grammar for Prompt Structure}

%The grammar representation for prompt structures is defined as follows:

\section{Results and Discussion}
\label{sec:results-discussion}

%This section presents the findings from our experiments, addressing three key perspectives: prompt diversity, feature correlation with performance, and prompt structure generalization. First, we evaluate the ability of MAP-Elites to enhance diversity in the phenotypic space compared to Random Search, focusing on the coverage of prompt structures. Next, we analyze correlations between specific phenotypic features (e.g., reasoning depth, number of examples) and performance to identify the characteristics most impactful for LLM effectiveness. Finally, we analyze the presence of specific prompt structures among the best-performing individuals to determine whether certain structures generalize well across multiple tasks or if task-specific tailoring is necessary for optimal performance. This analysis provides valuable insights into the relationship between prompt design and LLM effectiveness across diverse tasks.

This section presents findings from our experiments, focusing on three aspects: prompt diversity, feature-performance correlation, and prompt structure generalization. We compare MAP-Elites and Random Search in enhancing phenotypic diversity, analyze how features like reasoning depth and number of examples impact LLM performance, and examine whether top-performing prompt structures generalize across tasks or require task-specific tailoring. These insights highlight the relationship between prompt design and LLM effectiveness.

\subsection{Quality-Diversity Analysis}

%To analyze the diversity of prompts, we focused on two key phenotypic features: the number of examples (shots) and the depth of reasoning (chain of thought). These features were discretized into bins as described in the section \ref{sec:experiments}. To ensure the analysis emphasized both diversity and quality, only individuals achieving performance above 55\% were considered. For statistical robustness, we evaluated the quality-diversity coverage using three key tests. The Chi-Square Test measured the statistical difference between the distributions of MAP-Elites and Random Search, with higher values indicating greater disparity. The p-value assessed the significance of these differences, where values below 0.05 indicated statistical significance. Finally, Cramer’s V (Effect Size) quantified the strength of association between distributions, with values interpreted as small ($<$ 0.2), medium (0.2–0.5), or large ($>$ 0.5).

To analyze prompt diversity, we focused on two phenotypic features: the number of examples (shots) and reasoning depth (chain of thought), discretized into bins as described in Section \ref{sec:experiments}. We conducted the coverage analysis on all individuals but placed a focus on high performers (performance above 55\%) to emphasize quality. Then, we assessed quality-diversity coverage using three tests: the Chi-Square Test to compare MAP-Elites and Random Search distributions, the p-value to measure statistical significance ($<$ 0.05), and Cramer’s V (Effect Size) to quantify association strength (small: $<$ 0.2, medium: 0.2\text{–}0.5, large: $>$ 0.5).

\begin{table}[ht]
\centering
\caption{Prompt Phenotypic Space Coverage Results}
\resizebox{\columnwidth}{!}{%
\begin{tabular}{@{}lcccccc@{}}
\toprule
\textbf{Dataset} & \textbf{LLM} & \textbf{MAP (\%)} & \textbf{Random (\%)} & \textbf{Chi-Sq.} & \textbf{p-Val.} & \textbf{Effect Size} \\ \midrule
\cmidrule(lr){3-3} \cmidrule(lr){4-4}
 &  &HP\text{\textbar} Any & HP\text{\textbar} Any & &  &  \\ \midrule
\multirow{4}{*}{FFSN} 
 & LLaMA-3-1 & \textbf{72.0} \text{\textbar} \textbf{92.0} & 56.0 \text{\textbar} 80.0 & 0.78 & 0.3768 & 0.1250 \\
 & Qwen2-5   & \textbf{72.0} \text{\textbar} \textbf{92.0} & 44.0 \text{\textbar} 68.0 & 2.96 & 0.0856 & 0.2431 \\
 & Starling  & 56.0 \text{\textbar} \textbf{80.0} & 56.0 \text{\textbar} 68.0 & 0.00 & 1.0000 & 0.0000 \\
 & Phi-3.5   & \textbf{68.0} \text{\textbar} \textbf{92.0} & 60.0 \text{\textbar} 72.0 & 0.09 & 0.7683 & 0.0417 \\ \midrule
\multirow{4}{*}{KU} 
 & LLaMA-3-1 & \textbf{68.0} \text{\textbar}\textbf{100.0} & 48.0 \text{\textbar} 68.0  & 1.31 & 0.2517 & 0.1621 \\
 & Qwen2-5   & \textbf{56.0} \text{\textbar} 84.0 & 52.0 \text{\textbar} 84.0& 0.00 & 1.0000 & 0.0000 \\
 & Starling  & \textbf{68.0}  \text{\textbar} \textbf{96.0} & 52.0  \text{\textbar} 80.0 & 0.75 & 0.3865 & 0.1225 \\
 & Phi-3.5   & 64.0  \text{\textbar} 80.0 & \textbf{68.0}  \text{\textbar} \textbf{84.0}& 0.00 & 1.0000 & 0.0000 \\ \midrule
\multirow{4}{*}{LD3} 
 & LLaMA-3-1 & 20.0 \text{\textbar} \textbf{96.0} & 20.0 \text{\textbar} 72.0 & 0 & 1 & 0 \\
 & Qwen2-5   & 20.0 \text{\textbar}\textbf{100.0} & 20.0 \text{\textbar} 60.0 & 0 & 1 & 0 \\
 & Starling  & 20.0 \text{\textbar} \textbf{88.0}& 20.0  \text{\textbar} 64.0 & 0 & 1 & 0 \\
 & Phi-3.5   & 20.0 \text{\textbar} \textbf{84.0} & 20.0 \text{\textbar} 64.0 & 0 & 1 & 0 \\ \midrule
\multirow{4}{*}{PDSD} 
 & LLaMA-3-1 & 72.0 \text{\textbar} \textbf{88.0} & \textbf{80.0} \text{\textbar} 84.0 & 0.11 & 0.740 & 0.046 \\
 & Qwen2-5   & \textbf{80.0} \text{\textbar} \textbf{88.0} & 52.0 \text{\textbar} 52.0 & 3.21 & 0.073 & 0.253 \\
 & Starling  & \textbf{76.0} \text{\textbar} \textbf{92.0} & 56.0 \text{\textbar} 68.0  & 1.43 & 0.232 & 0.168 \\
 & Phi-3.5   & \textbf{76.0} \text{\textbar} \textbf{84.0} & 64.0 \text{\textbar} 80.0 & 0.38 & 0.537 & 0.087 \\ \midrule
\multirow{4}{*}{SSB} 
 & LLaMA-3-1 & \textbf{92.0} \text{\textbar} \textbf{92.0} & 64.0 \text{\textbar} 64.0 & 4.2 & \textbf{0.040} & \textbf{0.289} \\
 & Qwen2-5   & \textbf{96.0}  \text{\textbar} \textbf{96.0} & 60.0  \text{\textbar} 60.0 & 7.46 & \textbf{0.006} & \textbf{0.386} \\
 & Starling  & \textbf{84.0} \text{\textbar} \textbf{84.0} & 48.0 \text{\textbar} 48.0 & 5.7 & \textbf{0.016} & \textbf{0.337} \\
 & Phi-3.5   & \textbf{88.0} \text{\textbar} \textbf{88.0} & 68.0 \text{\textbar} 68.0 & 1.86 & 0.172 & 0.193 \\ \midrule
\multirow{4}{*}{SQA} 
 & LLaMA-3-1 & \textbf{72.0} \text{\textbar} \textbf{92.0} & 60.0 \text{\textbar} 72.0 & 0.36 & 0.550 & 0.008 \\
 & Qwen2-5   & \textbf{76.0} \text{\textbar} \textbf{96.0} & 48.0 \text{\textbar} 76.0 & 3.06 & 0.080 & 0.247 \\
 & Starling  & \textbf{76.0} \text{\textbar}\textbf{100.0} & 52.0 \text{\textbar} 68.0 & 2.17 & 0.140 & 0.208 \\
 & Phi-3.5   & \textbf{60.0} \text{\textbar} \textbf{92.0} & 52.0 \text{\textbar} 56.0 & 0.08 & 0.775 & 0.040 \\ \midrule
\multirow{4}{*}{Winowhy} 
 & LLaMA-3-1 & \textbf{88.0} \text{\textbar} \textbf{92.0} & 52.0 \text{\textbar} 76.0 & 6.1 & \textbf{0.0136} & \textbf{0.349} \\
 & Qwen2-5   & \textbf{96.0} \text{\textbar} \textbf{96.0} & 68.0 \text{\textbar} 76.0 & 4.88 & \textbf{0.0272} & \textbf{0.312} \\
 & Starling  & \textbf{96.0} \text{\textbar} \textbf{96.0} & 44.0 \text{\textbar} 48.0 & 13.71 &  \textbf{0.0002} & \textbf{0.523} \\
 & Phi-3.5   & \textbf{92.0} \text{\textbar} \textbf{92.0} & 56.0 \text{\textbar} 68.0 & 6.65 & \textbf{0.0099} & \textbf{0.364} \\ \bottomrule
\end{tabular}%
}
\label{tab:quality_diversity}
\end{table}

Table \ref{tab:quality_diversity} summarizes coverage results, where High Performers (HP) refers to individuals with performance above 55\%, and Any includes all individuals. Larger coverage values are bolded, and metrics are highlighted when p-values are below 0.05 with a medium/large effect size.

MAP-Elites consistently outperformed Random Search, achieving statistical significance in two tasks and covering over 60\% of the feature space with high-quality individuals in 21 of 28 runs, compared to 6 runs for Random Search.

%An interesting exception is observed with the LD3 task, which achieved high coverage rates when considering all individuals but covered only 20\% of the feature space with high-quality individuals. To investigate, we combined the archives from all four LLMs runs for LD3, selecting the best individuals, and visualized the results in Figure \ref{fig:LD-QD-chart}. Both MAP-Elites and Random Search show that only zero-shot prompts achieved performance above 60\%. This indicates that for the LD3 task, few-shot prompts may introduce logically consistent examples but also increase input complexity, making it harder for models to generalize or deduce correct logical conclusions. Thus, simpler zero-shot prompts appear to better align with the logical reasoning capabilities required for this task.

\begin{figure}
    \centering
    \includegraphics[width=1\linewidth]{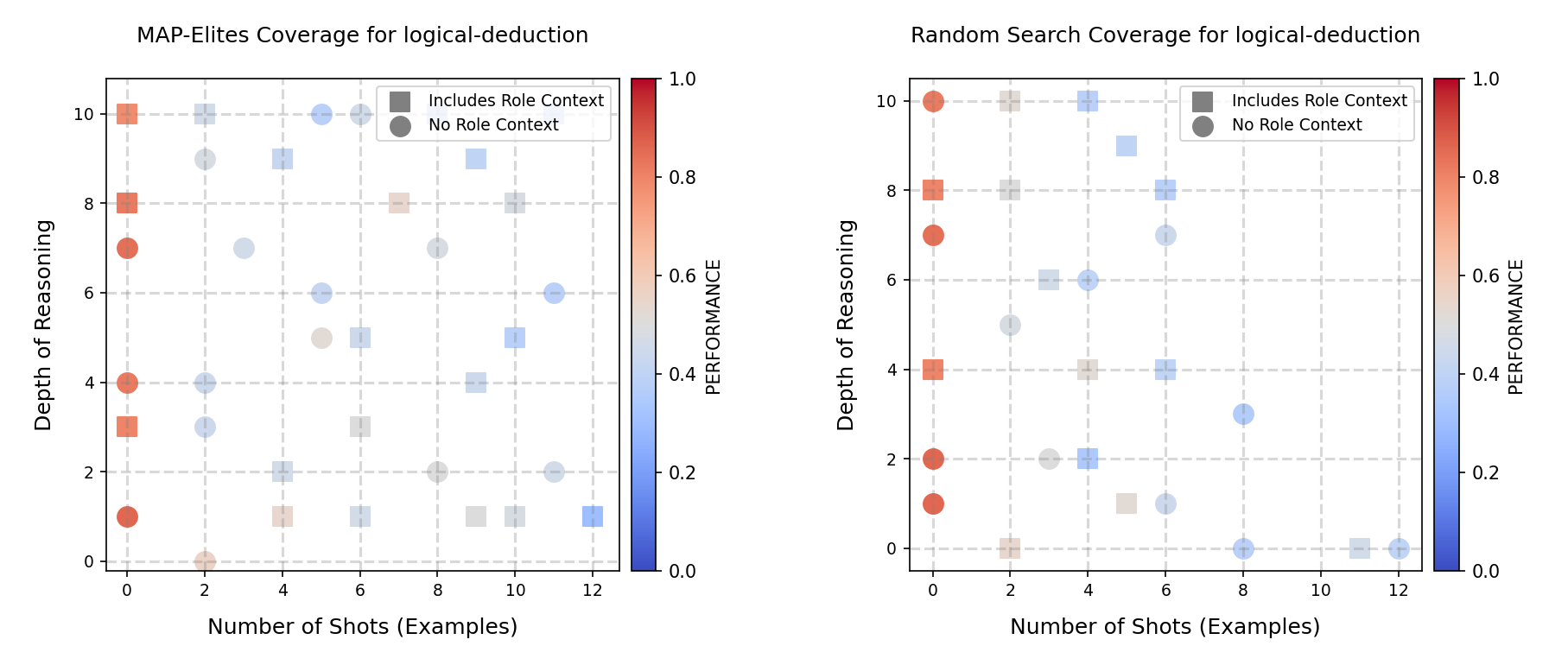}
    \caption{Feature space coverage for the Logical Deduction (LD3) dataset using MAP-Elites (left) and Random Search (right). The x-axis represents the number of examples, the y-axis denotes reasoning depth, and points indicate individuals, colored by performance (0.0–1.0) and shaped by role context inclusion. High performance was achieved only with zero-shot prompts, emphasizing the effectiveness of simpler designs.}
    \label{fig:LD-QD-chart}
\end{figure}

Interestingly, the LD3 task showed high coverage when considering all individuals but only 20\% with high-quality prompts. Combined archives from all LLM runs (Figure \ref{fig:LD-QD-chart}) reveal that only zero-shot prompts achieved high performance. 

This is probably due to the nature of logical deduction problems, where the selection of examples capable of accelerating the proof process requires a complex analysis of the morphology of the logical structures involved, which, in the case of evolutionary or random processes, would demand an impractical execution time to produce individuals containing suitable examples. It also justifies the low performance of the individuals with few-shot examples.

%This suggests that while few-shot prompts provide logically consistent examples, they may increase input complexity, hindering models’ ability to generalize. Simpler zero-shot prompts aligned better with the logical reasoning demands of LD3 in our experiments.

\subsection{Performance and Phenotypic Analysis}

%To analyze the relationship between prompt structures and LLM performance, we utilized all data generated from both MAP-Elites and Random Sampling approaches. This included the full set of created prompts, their extracted phenotypic features, and the corresponding model responses. For this analysis, we computed the Spearman correlation coefficient, which measures monotonic relationships between variables and is robust to non-linear dependencies. Statistical significance was assessed using p-values, providing insight into the strength and reliability of the observed correlations.

We analyzed the relationship between prompt structures and LLM performance using data from MAP-Elites and Random Sampling. Spearman correlation coefficients measured monotonic relationships, with p-values assessing the strength and significance of the correlations.

The analysis also included the type-token ratio, which measures lexical diversity by dividing the number of unique occurrence words (types) by the total word count (tokens), providing insights into the linguistic variability of the prompts.

\begin{figure}
    \centering
    \includegraphics[width=1\linewidth]{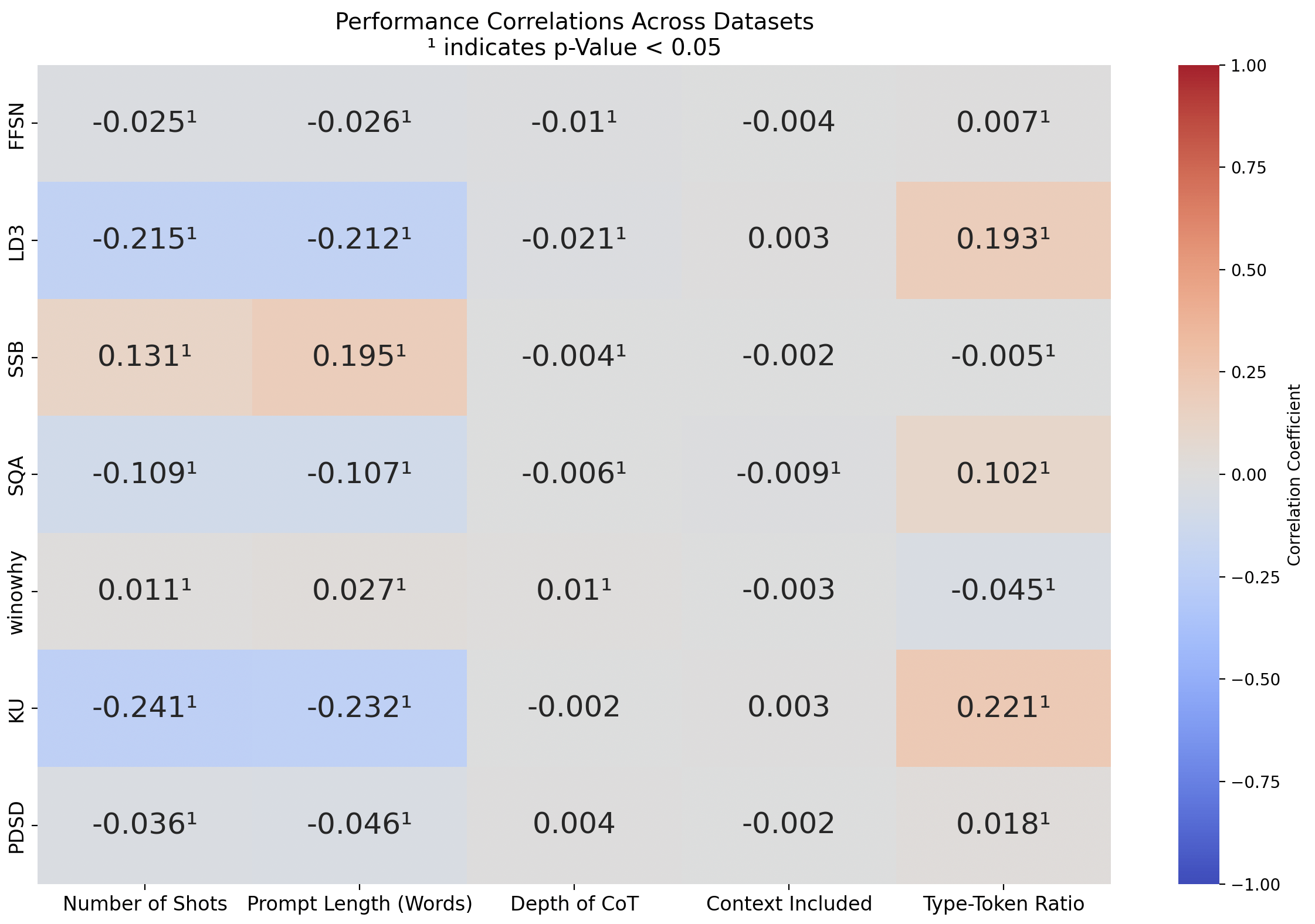}
    \caption{Heatmap showing correlations between prompt features and LLM performance across datasets. Significant correlations (p $<$ 0.05) are marked with $^{1}$.}
    \label{fig:correlation-table}
\end{figure}

The correlation analysis (Figure \ref{fig:correlation-table}) provides task-specific insights into how prompt features influence LLM performance. For tasks like LD3 and KU, modest negative correlations were observed between performance and features such as the number of shots and prompt length ( r $\approx$ -0.21 \text{ to } -0.24, p $<$ 0.05 ), suggesting that simpler prompts may reduce complexity and enhance accuracy. In contrast, SSB and SQA showed weak positive correlations with prompt length and type-token ratio ( r $>$ 0.1, p $<$ 0.05 ), indicating that richer, more diverse inputs could benefit narrative-based tasks.

However, the correlations were generally weak, limiting their practical significance. Negligible correlations for reasoning depth and context inclusion ( r $\approx$ 0 ) suggest minimal impact from these features across datasets. Some statistical significance may stem from large sample sizes rather than meaningful effects.

Despite these limitations, the analysis reveals variability in feature importance across tasks, emphasizing the need for task-specific prompt optimization.

\subsection{Evaluating Top-Performing Prompts}

To evaluate the effectiveness of prompt structures, we analyzed high performers from the MAP-Elites runs, calculating the percentage exhibiting features like examples (zero-shot, few-shot, many-shot), reasoning depth, and context. We compared this subset to the overall population of ~500 individuals generated per run, establishing a baseline for diversity.

\begin{table*}[!ht]
\centering
\caption{Distribution of phenotypic characteristics across datasets and models during MAP-Elites runs. The table highlights the percentage of individuals with features like context inclusion, example types (0-shot, few-shot, many-shot), and reasoning depth (No CoT, CoT-1, CoT2+), showcasing exploration diversity per model and dataset.}
\label{tab:distribution}
\resizebox{\textwidth}{!}{%
\begin{tabular}{llccccccccc}
\toprule
\multirow{2}{*}{Dataset} & \multirow{2}{*}{Model} & Has Context & \multicolumn{3}{c}{Examples} & \multicolumn{3}{c}{Reasoning} \\
\cmidrule(r){4-6} \cmidrule(r){7-9}
 &  & (\%) & 0-shot (\%) & Few-shot (\%) & Many-shot (\%) & No CoT (\%) & CoT-1 (\%) & CoT2+ (\%) \\
\midrule

\multirow{4}{*}{FFSN} 
 & Starling & 50.6 & 26.2 & 38.2 & 35.6 & 27.2 & 41.2 & 31.6 \\
 & LLaMA-3-1 & 57.6 & 23.2 & 45.4 & 31.4 & 28.4 & 36.4 & 35.2 \\
 & Phi-3.5 & 55.5 & 24.4 & 39.9 & 35.7 & 35.1 & 35.1 & 29.8 \\
 & Qwen2-5 & 40.0 & 25.4 & 42.4 & 32.2 & 24.8 & 37.8 & 37.4 \\
\midrule

\multirow{4}{*}{LD3} 
 & Starling & 52.4 & 22.4 & 44.0 & 33.6 & 28.2 & 36.6 & 35.2 \\
 & LLaMA-3-1 & 51.2 & 26.6 & 40.4 & 33.0 & 27.6 & 33.4 & 39.0 \\
 & Phi-3.5 & 46.6 & 25.2 & 43.4 & 31.4 & 26.8 & 38.8 & 34.4 \\
 & Qwen2-5 & 55.6 & 25.2 & 38.8 & 36.0 & 28.0 & 39.2 & 32.8 \\
\midrule

\multirow{4}{*}{SSB} 
 & Starling & 46.8 & 25.8 & 42.6 & 31.6 & 29.4 & 37.8 & 32.8 \\
 & LLaMA-3-1 & 48.4 & 24.2 & 41.4 & 34.4 & 26.4 & 37.2 & 36.4 \\
 & Phi-3.5 & 49.2 & 25.0 & 44.4 & 30.6 & 22.8 & 45.8 & 31.4 \\
 & Qwen2-5 & 50.4 & 25.0 & 45.4 & 29.6 & 26.4 & 36.2 & 37.4 \\
\midrule

\multirow{4}{*}{SQA} 
 & Starling & 50.5 & 27.3 & 40.8 & 31.9 & 30.0 & 37.2 & 32.7 \\
 & LLaMA-3-1 & 55.4 & 24.0 & 38.4 & 37.6 & 28.2 & 35.2 & 36.6 \\
 & Phi-3.5 & 58.4 & 27.8 & 38.0 & 34.2 & 21.4 & 40.6 & 38.0 \\
 & Qwen2-5 & 58.2 & 27.8 & 36.8 & 35.4 & 27.4 & 37.6 & 35.0 \\
\midrule

\multirow{4}{*}{PDSD} 
 & Starling & 56.3 & 23.5 & 49.9 & 26.6 & 20.5 & 40.7 & 38.9 \\
 & LLaMA-3-1 & 55.0 & 24.6 & 48.7 & 26.8 & 21.9 & 37.0 & 41.1 \\
 & Phi-3.5 & 58.2 & 21.1 & 52.3 & 26.5 & 26.8 & 40.2 & 33.0 \\
 & Qwen2-5 & 61.0 & 20.6 & 45.9 & 33.5 & 30.5 & 31.8 & 37.7 \\
\midrule

\multirow{4}{*}{KU} 
 & Starling & 53.8 & 24.8 & 39.4 & 35.8 & 28.4 & 37.6 & 34.0 \\
 & LLaMA-3-1 & 47.6 & 26.4 & 38.4 & 35.2 & 27.0 & 40.4 & 32.6 \\
 & Phi-3.5 & 46.0 & 26.0 & 40.4 & 33.6 & 27.7 & 39.9 & 32.5 \\
 & Qwen2-5 & 53.8 & 29.2 & 37.8 & 33.0 & 24.4 & 39.4 & 36.2 \\
\midrule

\multirow{4}{*}{Winowhy} 
 & Starling & 49.2 & 21.2 & 41.8 & 37.0 & 26.2 & 34.6 & 39.2 \\
 & LLaMA-3-1 & 48.6 & 23.2 & 40.6 & 36.2 & 28.0 & 35.6 & 36.4 \\
 & Phi-3.5 & 46.2 & 25.6 & 39.6 & 34.8 & 23.8 & 39.2 & 37.0 \\
 & Qwen2-5 & 50.6 & 25.6 & 38.0 & 36.4 & 27.2 & 39.0 & 33.8 \\
\bottomrule

\end{tabular}%
}
\end{table*}

Using a z-test for proportions, we tested whether specific features were disproportionately represented among high performers, with significant results indicating over- or under-representation. This analysis highlighted which prompt structures, such as zero-shot prompts or specific reasoning depths, contributed most to high performance, offering insights into effective designs for LLM tasks.

\begin{table*}[ht!]
\centering
\caption{The table highlights the distribution of features such as context inclusion, number of examples (0-shot, few-shot, many-shot), and reasoning depth (No CoT, CoT-1, CoT2+) among high performers. A dagger (†) indicates statistically significant differences with p-value $<$ 0.05.}
\label{tab:results_presence}
\resizebox{\textwidth}{!}{%
\begin{tabular}{llcccccccccc}
\toprule
\multirow{2}{*}{Dataset} & \multirow{2}{*}{Model} & \multirow{2}{*}{High Performers} & Has Context & \multicolumn{3}{c}{Examples} & \multicolumn{3}{c}{Reasoning} \\
\cmidrule(r){5-7} \cmidrule(r){8-10}
 &  &  & (\%) & 0-shot (\%) & Few-shot (\%) & Many-shot (\%) & No CoT (\%) & CoT-1 (\%) & CoT2+ (\%) \\
\midrule

\multirow{4}{*}{FFSN} 
 & Starling & 133 & 52.6 & 28.6 & 38.3 & 33.1 & 34.6 & 37.6 & 27.8 \\
 & LLaMA-3-1 & 132 & 63.6 & 20.5 & 50.8 & 28.8 & 24.2 & 41.7 & 34.1 \\
 & Phi-3.5 & 184 & 58.2 & 50.5\textsuperscript{\textdagger} & 28.3\textsuperscript{\textdagger} & 21.2\textsuperscript{\textdagger} & 37.0 & 34.8 & 28.3 \\
 & Qwen2-5 & 187 & 34.8 & 43.9\textsuperscript{\textdagger} & 31.0\textsuperscript{\textdagger} & 25.1 & 23.5 & 39 & 37.4 \\
\midrule

\multirow{4}{*}{LD3} 
 & Starling & 83 & 45.8 & 98.8\textsuperscript{\textdagger} & 1.2\textsuperscript{\textdagger} & 0.0\textsuperscript{\textdagger} & 31.3 & 34.9 & 33.7 \\
 & LLaMA-3-1  & 83 &  71.1\textsuperscript{\textdagger} & 100.0\textsuperscript{\textdagger} & 0.0\textsuperscript{\textdagger} & 0.0\textsuperscript{\textdagger} & 37.3 & 31.3 & 31.3 \\
 & Phi-3.5 & 97 & 40.2 & 99.0\textsuperscript{\textdagger} & 1.0\textsuperscript{\textdagger} & 0.0\textsuperscript{\textdagger} & 34.0 & 36.1 & 29.9 \\
 & Qwen2-5 & 123 & 54.5 & 100.0\textsuperscript{\textdagger} & 0.0\textsuperscript{\textdagger} & 0.0\textsuperscript{\textdagger} & 27.6 & 39.0 & 33.3 \\
\midrule

\multirow{4}{*}{SSB} 
 & Starling & 484 & 46.5 & 26.7 & 42.1 & 31.2 & 29.8 & 37.6 & 32.6 \\
 & LLaMA-3-1 & 474 & 48.1 & 25.5 & 40.7 & 33.8 & 25.9 & 37.8 & 36.3 \\
 & Phi-3.5 & 458 & 51.5 & 22.7 & 46.7 & 30.6 & 22.7 & 47.4 & 29.9 \\
 & Qwen2-5 & 475 & 50.1 & 26.4 & 44.8 & 28.8 & 26.5 & 36.4 & 37.1 \\
\midrule

\multirow{4}{*}{SQA} 
 & Starling & 288 & 50.7 & 55.9\textsuperscript{\textdagger} & 26.0\textsuperscript{\textdagger} & 18.1\textsuperscript{\textdagger} & 31.2 & 37.2 & 31.6 \\
 & LLaMA-3-1 & 236 & 53.4 & 48.3\textsuperscript{\textdagger} & 28.0\textsuperscript{\textdagger} & 23.7\textsuperscript{\textdagger} & 33.5 & 33.5 & 33.1 \\
 & Phi-3.5 & 220 & 55.0 & 44.5\textsuperscript{\textdagger} & 31.4 & 24.1\textsuperscript{\textdagger} & 25.0 & 36.8 & 38.2 \\
 & Qwen2-5 & 221 & 57.9 & 56.1\textsuperscript{\textdagger} & 24.0\textsuperscript{\textdagger} & 19.9\textsuperscript{\textdagger} & 29.0 & 33.9 & 37.1 \\
\midrule

\multirow{4}{*}{PDSD} 
 & Starling & 176 & 56.2 & 10.8 & 58.0 & 31.2 & 23.3 & 40.3 & 36.4 \\
 & LLaMA-3-1 & 252 & 54.4 & 33.3\textsuperscript{\textdagger} & 43.7 & 23.0 & 21.4 & 36.9 & 41.7 \\
 & Phi-3.5 & 239 & 58.2 & 28.9\textsuperscript{\textdagger} & 46.9 & 24.3 & 23.0 & 40.6 & 36.4 \\
 & Qwen2-5 & 263 & 59.7 & 27.0\textsuperscript{\textdagger} & 44.5 & 28.5 & 31.6 & 29.3 & 39.2 \\ 
\midrule

\multirow{4}{*}{KU} 
 & Starling & 200 & 50.5 & 62.0\textsuperscript{\textdagger} & 20.0\textsuperscript{\textdagger} & 18.0\textsuperscript{\textdagger} & 28.0 & 40.5 & 31.5 \\
 & LLaMA-3-1 & 226 & 44.7 & 58.4\textsuperscript{\textdagger} & 22.6\textsuperscript{\textdagger} & 19.0\textsuperscript{\textdagger} & 26.5 & 42.1 & 31.4 \\
 & Phi-3.5 & 260 & 51.5 & 60.8\textsuperscript{\textdagger} & 22.7\textsuperscript{\textdagger} & 16.5\textsuperscript{\textdagger} & 29.6 & 38.1 & 32.3 \\
 & Qwen2-5 & 210 & 51.4 & 69.5\textsuperscript{\textdagger} & 16.7\textsuperscript{\textdagger} & 13.8\textsuperscript{\textdagger} & 21.9 & 41.4 & 36.7 \\
\midrule

\multirow{4}{*}{Winowhy} 
 & Starling & 443 & 49.2 & 21.0 & 43.1 & 35.9 & 26.6 & 33.0 & 40.4 \\
 & LLaMA-3-1 & 286 & 45.8 & 18.5 & 43.0 & 38.5 & 25.5 & 34.3 & 40.2 \\
 & Phi-3.5 & 247 & 44.1 & 30.4 & 32.4 & 37.2 & 22.3 & 38.9 & 38.9 \\
 & Qwen2-5 & 440 & 50.0 & 25.2 & 37.3 & 37.5 & 26.4 & 38.9 & 34.8 
\\
\bottomrule

\end{tabular}%
}
\end{table*}

Table \ref{tab:distribution} shows the percentage distribution of phenotypic features (context inclusion, examples: 0-shot, few-shot, many-shot, and reasoning depth: No CoT, CoT-1, CoT2+) across all individuals involved on the MAP-Elites runs, highlighting the diversity of explored prompt structures.

Table \ref{tab:results_presence} reports the number of high-performing individuals and their feature distributions. Statistically significant differences (†, p $<$ 0.05) between high performers and the overall population reveal variability in feature prevalence across datasets and models.

\subsection{Discussion}

The results reveal that optimal prompt structures vary significantly depending on the type of task. For datasets such as PDSD, SSB, and Winowhy, which involve tasks with a more contextual and pattern-recognition nature, prompts with examples (e.g., few-shot or many-shot) performed better. This trend likely stems from the benefits of in-context learning \cite{min2022rethinking}, where the model uses provided examples to learn schemas, templates, and patterns that guide it toward accurate responses \cite{swaminathan2024schema}. For instance, in the Winowhy dataset, identifying pronoun references often relies on examples that highlight consistency between contextual elements and their resolution, allowing the model to generalize effectively within the scope of the task.

Conversely, example-based datasets did not provide good results for datasets such as LD3, SQA, and Known Unknowns, which involve tasks that are structurally diverse and strongly rely on external or implicit knowledge. For these tasks, zero-shot prompts consistently outperformed few-shot prompts, likely  because introducing examples increased task complexity, distracting the model from the core reasoning process. In these cases, zero-shot prompts seem to make the model rely on its pre-trained reasoning capabilities, without being constrained by a limited set of examples \cite{kojima2022large}. This reduction in cognitive load and avoidance of overfitting to narrow task representations likely enables better performance in tasks requiring generalized reasoning.

Concerning the contribution of the phenotype $context$, the experiments showed that such a feature statistically did not impact the performance of the overall investigated prompts. One possible reason for this outcome is the generic nature of the context-role templates used, which were pre-defined and constant across tasks. This approach may have limited alignment with task-specific nuances, suggesting that more dynamically generated or task-sensitive context roles might be necessary to uncover the true potential of this feature.

Likewise, chain-of-thought (CoT) reasoning structures did not demonstrate significant performance gains compared to their representation in the overall population. While CoT reasoning was prominent among high-performing individuals, it was not a universal determinant of success. 

%On the other hand, the phenotypic structures $chain-of-thought (CoT) reasoning$, despite not impacting the performance of the individuals of the overall population, were effective in contributing to increasing the performance of the high performing individuals. 

Noteworthy here is that, in some cases, zero-shot prompts without explicit reasoning slightly outperformed those with reasoning. This observation suggests that simpler prompts may align better with the reasoning capabilities of certain models, avoiding the additional cognitive overhead that comes with parsing and integrating complex reasoning chains.

Finally, the performed experiments, further than ratifying the effectiveness of the approach proposed herein to deal with automated prompt engineering, also answered the questions presented in the beginning of Section \ref{sec:experiments}.

By systematically exploring diverse prompt structures, this approach enables a deeper understanding of how structural variations influence performance across a range of tasks. The results highlight the importance of balancing simplicity and complexity in prompt design to optimize LLM performance, paving the way for more targeted and effective strategies in the future.

\section{Conclusion and Future Work}
\label{sec:conclusion}
This work demonstrates that combining CFG  with QD algorithms like MAP-Elites effectively enhances prompt diversity, significantly outperforming random generation. This method expands the prompt space and offers valuable insights into how structural variations influence LLM performance, supporting both LLM training and In-Context Learning techniques.

%Our findings reveal mild correlations between phenotypic characteristics—such as the number of examples, reasoning depth, and context inclusion—and task performance. While not strong, these correlations suggest that thoughtful prompt design can enhance performance in specific scenarios. 

Our findings reveal mild correlations between some phenotypic characteristics and task performance. However, even in situations in which such correlations are not high, they suggest that thoughtful prompt design can enhance performance in specific scenarios.

%However, optimal prompt structures vary across tasks; simpler prompts like zero-shot often excel in some tasks, while others benefit from more complex designs like few-shot or reasoning-intensive prompts. This task dependency highlights the value of aligning prompt design with task characteristics, a process MAP-Elites facilitates by systematically generating diverse, high-performing prompts.

The experiments showed that the effectiveness of the prompt structures depends on the nature of the tasks. In fact, simpler prompts like zero-shot often excel in some tasks, while others benefit from more complex designs like few-shot or reasoning-intensive prompts. The experiments also proved that the approach proposed in this work succeeds in generating high performing individuals able to deal with the specificities associated with the various tasks.

Future research should explore broader feature spaces to deepen the understanding of QD algorithms in relation to In-Context Learning, including real-world applications to validate practical utility. Combining QD-based methods with traditional optimization techniques like Reinforcement Learning or Gradient-Based approaches could further enhance results, leveraging QD’s diversity alongside the precision of traditional methods to advance prompt engineering and task performance.

\subsection{Limitations}

This study faced resource-related limitations. Experiments were restricted to seven tasks and four LLMs (under 10B parameters) due to cost constraints—around \$500 was spent on Hugging Face. Results may not generalize to larger models. All experiments ran on a single MacBook Pro (Apple M3 Max), limiting iteration counts, population size, and algorithm diversity. Internet-based inferences also introduced latency. Future work could overcome these barriers with improved hardware and local execution for greater scalability and depth.
\bibliography{main.bib}

\vspace{12pt}

\end{document}